\documentclass[12pt]{article}
\usepackage{amsmath}
\usepackage{graphicx,psfrag,epsf,color,subfigure}
\usepackage{enumerate}
\usepackage{natbib}
\usepackage[unicode=true,pdfusetitle,
   bookmarks=true, 
   bookmarksnumbered=false, 
   bookmarksopen=false,
   breaklinks=false, 
   pdfborder={0 0 1}, 
   backref=false, 
   colorlinks=false]{hyperref}

\usepackage{amssymb}
\usepackage{multirow}
\usepackage{array}
\usepackage{bm}
\usepackage{booktabs}
\usepackage{float}
\usepackage[font=small,labelfont=bf]{caption}
\setcitestyle{citesep={;}}
\usepackage[total={6.4in,8.6in}]{geometry}
\usepackage{algorithm}
\usepackage{algorithmic}
\usepackage{amsmath}
\usepackage{amssymb}
\usepackage{mathtools}
\usepackage{amsthm}
\usepackage{graphicx,psfrag,epsf,color}
\usepackage{enumerate}
\usepackage{natbib}
\usepackage{enumitem}
\usepackage{multirow}
\usepackage{array}
\usepackage{bm}
\usepackage{booktabs}
\usepackage{float}  
\usepackage[capitalize,noabbrev]{cleveref}

\theoremstyle{plain}
\newtheorem{theorem}{Theorem}

\newtheorem{lemma}{Lemma}

\theoremstyle{definition}

\newtheorem{assumption}{Assumption}
\theoremstyle{remark}

\usepackage[textsize=tiny]{todonotes}

\newcommand{\emptycomment}[1]{}
\newcommand{\indep}{\;\, \rule[0em]{.03em}{.67em} \hspace{-.27em}
	\rule[-.02em]{.7em}{.03em} \hspace{-.27em}
	\rule[0em]{.03em}{.67em}\;\,}
\newcommand{\bse}{\begin{eqnarray*}}
\newcommand{\ese}{\end{eqnarray*}}

\usepackage{authblk}

\begin{document}

\def\spacingset#1{\renewcommand{\baselinestretch}%
{#1}\small\normalsize} \spacingset{1}
%
%
%%%%%%%%%%%%%%%%%%%%%%%%%%%%%%%%%%%%%%%%%%%%%%%%%%%%%%%%%%%%%%%%%%%%%%%%%%%%%%%
\title{\bf \Large Conformal Diffusion Models for Individual Treatment Effect Estimation and Inference}
\author[1]{Hengrui Cai\thanks{Equal contribution.}}
\author[2]{Huaqing Jin$^*$} 
\author[3]{Lexin Li}
\affil[1]{\small Department of Statistics, University of California Irvine} 
\affil[2]{Department of Epidemiology and Biostatistics, 
  University of California San Francisco} 
\affil[3]{Department of Biostatistics and Epidemiology, University of California Berkeley} 

 \date{}
 
 \maketitle  

\baselineskip=21pt

\begin{abstract}
Estimating treatment effects from observational data is of central interest across numerous application domains. Individual treatment effect offers the most granular measure of treatment effect on an individual level, and is the most useful to facilitate personalized care. However, its estimation and inference remain underdeveloped due to several challenges. In this article, we propose a novel conformal diffusion model-based approach that addresses those intricate challenges. We integrate the highly flexible diffusion modeling, the model-free statistical inference paradigm of conformal inference, along with propensity score and covariate local approximation that tackle distributional shifts. We unbiasedly estimate the distributions of potential outcomes for individual treatment effect, construct an informative confidence interval, and establish rigorous theoretical guarantees. We demonstrate the competitive performance of the proposed method over existing solutions through extensive numerical studies. 
\end{abstract}

%%%%%%%%%%%%%%%%%%%%%%%%%%%%%%%%%%%%%%%%%%%%%%%%%%%%%%%%%%%%%%%%%%%%%%%%%%%%%%%%%%%%%%%%%%%%%%%%
%\vspace{-0.15cm}
\section{Introduction} \label{sec:intro}
%\vspace{-0.1cm}
Estimating treatment effects from observational data is of central interest across a wide range of applications, spanning public health \citep{cole2020impact, goodman2020using}, political science \citep{abadie2010synthetic, sabia2012effects}, economics \citep{cavallo2013catastrophic, dube2015pooling}, and beyond. Over the past decades, the extensive literature has focused on estimating the average treatment effect (ATE), which provides a measure of the average effectiveness of a treatment for a population of subjects \citep[see, for example,][]{rubin1978bayesian, pearl2009causality}. More recently, some literature has shifted the interest to estimating the conditional average treatment effect (CATE), which provides a more refined measure of the average treatment effect for subgroups of the population defined by specific characteristics or covariates \citep[see, for example,][]{athey2015machine, shalit2017estimating, wager2018estimation, kunzel2019metalearners, farrell2021deep}. While CATE offers a more detailed and informative understanding of the treatment effect than ATE, both measures overlook the inherent variability in individual responses to treatment, under which the effect of treatment can vary significantly between individuals. Such subject variability can be crucial for decision-making. For instance, while a treatment may prove effective for 90\% of patients aged 60 or older, its efficacy for a specific individual may still remain uncertain. As such, there has been growing interest in estimating the individual treatment effect (ITE), which quantifies the effect of the treatment on an individual level \citep{wager2018estimation,lei2021conformal,alaa2023conformal}. Such a measure seeks the most granular insight into treatment effects and is particularly useful in facilitating personalized care. 

Despite the clear importance and demand for a better understanding of ITE, its estimation and inference remain underdeveloped, beset by some major challenges. First, inherent in the definition of ITE is the constraint that each individual can only be treated once, and ITE is a random quantity rather than a population parameter such as ATE or CATE. These characteristics pose challenges when working with observational data. Current solutions for ITE often rely on relatively simple models, or some strong distributional assumptions on the noise term, failing to adequately capture the complex treatment effect. Second, the presence of confounding variables often results in the covariate distributional shift, manifesting as differing distributions of covariates between the calibration and testing datasets, as well as between the treated and control groups. This covariate shift can lead to model inaccuracies and poor generalization performance. Consequently, there is a pressing need to develop new approaches to both flexibly capture complex treatment relations and effectively handle distributional shifts while maintaining theoretical guarantees. 

In this article, we propose a novel conformal diffusion model that addresses the intricate challenges of individual treatment effect estimation and inference in observational studies. Leveraging the flexibility and robustness of diffusion modeling \citep{sohl2015deep,ho2020denoising} and the model-free statistical inference paradigm of conformal inference \citep{lei2018distribution}, our method unbiasedly estimates the distributions of potential outcomes for individual treatment effect, and constructs an informative confidence interval with rigorous guarantees. Our contributions are multifold.

%\vspace*{-5pt}
\begin{itemize}[leftmargin=*]
\item To the best of our knowledge, we present the first diffusion model-based conformal inference solution for ITE. Toward this goal, we generalize the probabilistic conformal prediction, extend the conditional random sampling from the supervised regression setting  \citep{pmlr-v206-wang23n} to the ITE setting, and integrate with diffusion modeling. The resulting solution provides a highly flexible and effective confidence interval for ITE. 
%\vspace*{-2pt}
\item We integrate the propensity score and localization into our non-conformity score that effectively handles various types of covariate distributional shifts inherent in observational studies. More specifically, by incorporating the propensity score, we mitigate the covariate shift arising from potential assignment imbalance between the treated and control groups. By leveraging the local approximation of covariate information, we further improve the sample efficiency owing to the covariate shift between the calibration and testing samples.  
%\vspace*{-2pt}
\item We establish rigorous theoretical guarantees for the coverage of our proposed confidence interval for ITE under common regularity conditions. We demonstrate the superior performance of the proposed method over existing solutions through extensive numerical studies. 
\end{itemize}
%\vspace*{-5pt}

%%%%%%%%%%%%%%%%%%%%%%%%%%%%%%%%%%%%%%%%%%%%%%%%%%%%%%%%%%%%%%%%%%%%%%%%%%%%%%%%%%%%%%%%%%%%%%%%
%\vspace{-0.15cm}
\subsection{Related works}
%\vspace{-0.15cm}
\textbf{Individual treatment effect} (ITE) has gained considerable attention in numerous domains, such as political science \citep{imai2011treatment, grimmer2017understanding}, psychology \citep{bolger2019individual, winkelbeiner2019individual}, sociology \citep{xie2012causal, breen2015causal}, economics \citep{florens2008causal, djebbari2008identifying}, and education \citep{morgan2001counterfactual, brand2012causal}. ITE focuses on the effect of treatment at an individual level, contrasts traditional methods that focus on the effects of entire populations, and is particularly important for applications where the treatment responses vary considerably for individuals \citep{tian2000modeling, hernan2010causal}. Advanced machine learning techniques, especially those employing causal inference and predictive modeling, have been pivotal in refining the accuracy of ITE estimation and inference \citep{shalit2017estimating,alaa2018bayesian, louizos2017causal}. Techniques such as targeted maximum likelihood estimation \citep{van2011targeted} and machine learning-based propensity score matching \citep{stuart2010matching} provide additional rigor. Recent developments have started to incorporate deep learning techniques, enabling the capture of complex, nonlinear relations within data, thereby offering more detailed and individualized treatment effect estimation \citep{johansson2016learning,kunzel2019metalearners, risteski2020machine}. These advancements not only improve clinical decision-making, but also contribute to the burgeoning field of precision medicine, where treatments can be tailored to individual patients \citep{collins2015new, jameson2015precision}.

\textbf{Conformal inference} offers a robust statistical framework for assessing the uncertainty in predictive models, and has gained widespread applications in areas such as bioinformatics, financial modeling, and climate science \citep{vovk2005algorithmic, shafer2008tutorial, balasubramanian2014conformal, zeni2020conformal}. Its key strength lies in providing distribution-free confidence interval for prediction, making it particularly useful in fields requiring precise uncertainty quantification \citep{lei2018distribution}. Moreover, the versatility of conformal inference extends its compatibility with complex models, including deep neural networks. This compatibility allows for the integration of advanced deep learning techniques with robust statistical guarantees \citep{papadopoulos2008inductive, lei2018distribution}. Recent advancements have further integrated it with nonparametric methods such as Bayesian approaches, thereby enhancing its practicality and scope of application \citep{burnaev2016conformal, romano2020classification}. 

\textbf{Diffusion models} are a novel class of generative models, and have emerged as a powerful tool in areas such as image and speech generation \citep{sohl2015deep, ho2020denoising}. These models simulate the gradual transformation of noise into structured data distributions, mimicking physical diffusion processes. Notably, they excel in generating diverse, high-fidelity outputs, often surpassing the capabilities of traditional generative models such as GANs  \citep{dhariwal2021diffusion, nichol2021improved}. Their ability to generate detailed and varied outputs has made them a prominent choice for tasks requiring high levels of creativity and precision. Moreover, the versatility of diffusion models extends beyond media generation, and has found significant applications in fields like drug discovery and material design, where the ability to explore a vast space of molecular configurations is crucial \citep{xie2021crystal, yang2021diffusion}. Recent advancements have further enhanced the applicability of diffusion models, by combining diffusion process with other generative approaches \citep{kingma2021variational, vahdat2021score}.

%%%%%%%%%%%%%%%%%%%%%%%%%%%%%%%%%%%%%%%%%%%%%%%%%%%%%%%%%%%%%%%%%%%%%%%%%%%%%%%%%%%%%%%%%%%%%%%%
%\vspace{-0.15cm}
\section{Problem setup}\label{sec:setup}
%\vspace{-0.1cm}
%%%%%%%%%%%%%%%%%%%%%%%%%%%%%%%%%%%%%%%%%%%%%%%%%%%%%%%%%%%%%%%%%%%%%%%%%%%%%%%%%%%%%%%%%%%%%%%%
\subsection{Individual treatment effects}
%\vspace{-0.1cm}
We adopt the potential outcome framework of \citet{rubin1974estimating} throughout this article. Consider a sample of $n$ units, where each unit $i$ is associated with a $d$-dimensional covariate $X_i \in \mathcal{X} \subset  \mathbb{R}^d$, a binary treatment $T_i\in \{0,1\}$, and the potential outcomes $[Y_i(0), Y_i(1)]$. We assume the potential outcomes and treatment assignment are independent and identically distributed, i.e.,
$
\left(Y_i(0), Y_i(1), T_i, X_i\right) \stackrel{i . i . d .}{\sim} (Y(0), Y(1), T, X). $ 
The individual treatment effect (ITE) $\tau_i$ is defined as the difference between the potential outcomes under treatment and control conditions:
\begin{equation*}
\tau_i \equiv Y_i(1)-Y_i(0).
\end{equation*}
In practice, we only observe the factual outcome associated with the received treatment, i.e., $Y_i = Y\left(T_i\right)=T_i Y_i(1)+\left(1-T_i\right) Y_i(0)$. Thus, our observed dataset consists of the triplets $(Y_i, T_i, X_i)_{i=1}^n$. Since $\tau_i = Y_i -Y_i(0)$ if $T_i=1$ or $\tau_i = Y_i(1) -Y_i$ if $T_i=0$, the ITE $\tau_i$ is a random variable and cannot be estimated as a point value. Hence, our objective is to construct a confidence interval for the individual treatment effect $\tau_i$ that is tight and has the desired coverage property.

As each individual can only be treated once, $\tau_i$ is not estimable from the observational data,  unless some additional conditions are introduced. We postulate that the potential outcomes follow a general generating procedure as, $
Y(T)=f_T\left(X, E_T\right), $ 
where $f_T\left(X, E_T\right)$ represents an unknown function of the covariates $X$ and noise $E_T$ under different treatments $T\in \{0,1\}$. We remark that $f_T$ can be highly flexible, and we do not impose any particular parametric form for $f_T$. Moreover, we introduce the following conditions. 

\begin{assumption}\label{cons}
(Consistency): The observed outcome given a specific treatment equals its potential outcome, i.e., $Y=Y(T)$.
\end{assumption}
    
\begin{assumption}\label{ign}
(Ignorability): The potential outcomes are jointly independent of the treatment assignment conditional on $X$, in that $[Y(0), Y(1)] \indep T \mid X$.
\end{assumption}

\begin{assumption}\label{posi}
(Positivity): For all $X \in \mathcal{X} \subset \mathbb{R}^d$, the probability of receiving any treatment is bounded away from zero, in that $0 < c_0 < \operatorname{Pr}(T=1 \mid X=x) < c_1 < 1$ for some constants $c_0$ and $c_1$ for all $x$ that its density $P_X(x)>0$.
\end{assumption}

Assumption \ref{cons} ensures that the observed outcomes are consistent with the potential outcomes corresponding to the observed treatments. Assumption \ref{ign} requires that, given the baseline covariates $X$, it contains enough observed confounders to render the treatment assignment conditionally independent of the potential outcomes, which is crucial for addressing confounding bias in observational studies. Assumption \ref{posi} guarantees that there are no regions of the covariate space where the probability of receiving treatment is too close to zero or one, ensuring that the treatment effect can be estimated across the entire covariate space. All these conditions are commonly imposed in the causal inference literature \citep[see e.g.,][]{chen2016personalized,zhu2020kernel,schulz2020doubly,lei2021conformal}.

%%%%%%%%%%%%%%%%%%%%%%%%%%%%%%%%%%%%%%%%%%%%%%%%%%%%%%%%%%%%%%%%%%%%%%%%%%%%%%%%%%%%%%%%%%%%%%%%
%\vspace{-0.15cm}
\subsection{Conformal inference with sample splitting}
%\vspace{-0.1cm}
We briefly review the classic conformal inference with sample splitting. Specifically, we divide the observed triplets $(Y_i, T_i, X_i)_{i=1}^n$ into three parts: $\mathcal{D}_{\text{train}}={(X_i, T_i, Y_i)}_{i=1}^{n_{\text{train}}}$ as the training data, $\mathcal{D}_{\text{cal}}={(X_j, T_j, Y_j)}_{j=1}^{n_{\text{cal}}}$ as the calibration data, and $\mathcal{D}_{\text{test}}={(X_k, T_k, Y_k)}_{k=1}^{n_{\text{test}}}$ as the holdout testing or target data. Let $\mathcal{I}_{\text{train}}, \mathcal{I}_{\text{cal}}, \mathcal{I}_{\text{test}}$ denote the corresponding index set, respectively. Without loss of generality, we follow the general literature of ITE \citep[see e.g., ][]{lei2021conformal} and let $Y(0) = 0$. Then the confidence interval for $\tau_i$  is equivalent to that for $Y_i(1)$. Define
$\pi(x) = \mathbb{P}(T=1 \mid X=x)$ as the propensity score function that describes the treatment assignment mechanism, and $\mu(x) = \mathbb{E}[Y \mid X=x, T=1]$ as the expectation of the unknown complex function $\mathbb{E}[f_1\left(X, E_1\right)]$ over the noise under different treatment groups.  We first train the model $\widehat{\mu}(x)$ using $\mathcal{D}_{\text{train}}$, and compute the non-conformity score using  $\mathcal{D}_{\text{cal}}$ as $\mathcal{V}_j(\widehat{\mu}) = V\left(X_j, T_j, Y_j; \widehat{\mu}\right), \forall j \in \mathcal{I}_{\text{cal}}$, where $V(x, t, y ; \widehat{\mu}) = |\widehat{\mu}(x)-y|$. Then, for a given coverage level of $1-\alpha$, we compute the empirical quantile of the non-conformity scores for the treated unit as,
% %\vspace{-0.05in}
\begin{equation*}
Q_{\mathcal{V},1}(1-\alpha) = \text { Quantile }_{1-\alpha}\left(\sum_{j \in \mathcal{I}_{\text{cal}}}  \delta_{\mathcal{V}_j}+  \delta_{+\infty}\right),
\end{equation*}
where $\delta_s$ is the point mass at $s$. Finally, we construct the predictive interval for $Y_{k}$ at a new target point $X_{k}=x$ in $\mathcal{D}_{\text{test}}$ as, 
\begin{equation*}
\widetilde{C}(x)=\widehat{\mu}(x) \pm Q_{\mathcal{V},1}(1-\alpha),
\end{equation*}
where the predicted conditional mean at point $x$ serves as the mean value under the treated group, and the non-conformity score $Q_{\mathcal{V},1}$ serves as the quantification of variability of the individual response. When the calibration data and the testing data are fully exchangeable, this interval is guaranteed to satisfy the marginal coverage probability that $\mathbb{P}\left(Y_{k}(1) \in \widetilde{C}\left(X_{k}\right)\right) \geq 1-\alpha$ \citep{tibshirani2019conformal,kuchibhotla2020exchangeability,pmlr-v206-wang23n}.

%%%%%%%%%%%%%%%%%%%%%%%%%%%%%%%%%%%%%%%%%%%%%%%%%%%%%%%%%%%%%%%%%%%%%%%%%%%%%%%%%%%%%%%%%%%%%%%%
%\vspace{-0.15cm}
\section{Conformal diffusion models for ITE}\label{sec:method}
%\vspace{-0.15cm}
While conformal inference has been utilized to construct a confidence interval for the individual treatment effect $\tau_i$, existing approaches rely on some relatively simple models such as quantile regression or pseudo-outcome regression \citep{lei2021conformal,alaa2023conformal}, which limits their ability to handle complex treatment effects in real-world applications. Moreover, they usually require the exchangeability between the calibration data and the testing data, and do not explicitly tackle distributional shifts \citep{pmlr-v206-wang23n}. These observations have motivated us to develop a new and powerful approach, by integrating a highly flexible deep generative model, the conformal inference paradigm, along with the propensity score and covariate local approximation, for the estimation and inference of the individual treatment effect. Our proposed method comprises a few key components, which we elaborate on next.

%%%%%%%%%%%%%%%%%%%%%%%%%%%%%%%%%%%%%%%%%%%%%%%%%%%%%%%%%%%%%%%%%%%%%%%%%%%%%%%%%%%%%%%%%%%%%%%%
%\vspace{-0.15cm}
\subsection{Modeling potential outcome distribution via deep generative learning}
\label{step1}
%\vspace{-0.15cm}
We propose to use a deep generative model to learn the conditional distribution of the potential outcome given the covariates, i.e., $P(Y(T) \mid X)$, then compute the mean of the random samples from the learnt distribution. This departs from the existing solutions such as \citet{lei2021conformal} that rely on a point estimate of the conditional mean function $\mu$. By leveraging the expressive power of deep generative modeling, our method is expected to better capture the intricately complicated relations between the covariates and the potential outcome, to obtain a better uncertainty quantification, and subsequently to facilitate the ITE confidence interval construction. 

More specifically, we first build two separate diffusion models using the training data $\mathcal{D}_{\text{train}}$ to learn the conditional distributions $q_0(Y\mid X)$ and $q_1(Y\mid X)$ for $T=0$ and $T=1$, respectively. Denote the corresponding diffusion models as $\widehat{q}_0(Y\mid X)$ and $\widehat{q}_1(Y\mid X)$, respectively. Many versions of deep diffusion models can be utilized for this step, and we implement the denoising diffusion probabilistic model (DDPM) of \citet{song2020denoising}, which incorporates a probabilistic framework and enables a principled uncertainty quantification.  

We next compute the non-conformity scores based on the estimated diffusion models. Toward that end, we generalize the probabilistic conformal prediction and extend the conditional random sampling for supervised regression \citep{pmlr-v206-wang23n} to the setting of ITE and deep generative modeling. Taking $Y(1)$ as an example, for each $(X_j, T_j, Y_j) \in \mathcal{D}_{\text{cal}}$, we randomly sample $M$ samples,  $\widehat{Y}_{j1}, \ldots, \widehat{Y}_{jM} \sim \widehat{q}_{1}(Y|X_j)$. We calculate the non-conformity score as, 
\begin{eqnarray*}
V_j=\min_{1\le m \le M}|Y_j- \widehat{Y}_{jm}|.
\end{eqnarray*} 
Following the convention of the current literature \citep[see e.g., ][]{lei2021conformal}, we illustrate the construction of the confidence interval for the treated units only, whereas the inference for the controlled units can be established in a similar way.

%%%%%%%%%%%%%%%%%%%%%%%%%%%%%%%%%%%%%%%%%%%%%%%%%%%%%%%%%%%%%%%%%%%%%%%%%%%%%%%%%%%%%%%%%%%%%%%%
%\vspace{-0.15cm}
\subsection{Adjusting covariate distributional shifts}
\label{step2}
%\vspace{-0.1cm}
Directly using the mean difference between two separate deep generative models to construct the confidence interval for ITE may suffer from covariate distributional shift. This is because, the joint distribution of the observed samples $(X, Y)$ under the treated group $(T=1)$ in the calibration data is given by $P_{X \mid T=1} \times P_{Y(1) \mid X}$, while for the testing or target data, we have the corresponding joint distribution as $Q_X \times P_{Y(1) \mid X}$, where $Q_X$ is the covariate distribution in the target population. These two distributions share the same conditional distribution $P_{Y(1) \mid X}$ of the outcome, but otherwise differ in the distribution of the covariates. Moreover, in the control group, $X$ follows $P_{X \mid T=0}$, and in the treatment group, $X$ follows $P_{X \mid T=1}$, where we often have \(P_{X \mid T=0} \neq P_{X \mid T=1}\) in the observational studies. This leads to another covariate shift owing to treatment assignment. 
 
To adjust for the first potential distributional shift between the calibration data and the test data, we utilize the local approximation idea \citep{tibshirani2019conformal,guan2023localized,hore2023conformal}, which measures the covariate similarity between the calibration data and the target one, then reweighs the non-conformity scores based on such similarity. Toward that end, we utilize a certain kernel function to characterize the distribution of covariates in the calibration data and measure the similarity to a given testing or target data point $X_k$ of interest. We then reweigh the non-conformity scores in $\mathcal{D}_{\text {cal}}$ by placing more weight on $\{X_j\}_{j \in \mathcal{I}_{\text {cal}}}$ that lie near $X_{k}$. We choose the localization kernel \citep{guan2023localized} given by a kernel function $H: \mathcal{X} \times \mathcal{X} \rightarrow \mathbb{R}_{\geq 0}$, and we calculate the local weight as, 
\begin{eqnarray}\label{kernel}
\widehat{w}_{1}(X)=\frac{H\left(X, \widetilde{X}_{k}\right)}{\sum_{j \in \mathcal{I}_{\rm cal}\cup\{k\}} H\left(X_j, \widetilde{X}_{k}\right)}, 
\end{eqnarray}
%{\color{blue} 
where $\widetilde{X}_{k}$ is sampled from the density function proportional to $H(X_{k}, \cdot)$. We note that the weights in\eqref{kernel} utilize the local covariate information in the calibration data, and thus improving the sample efficiency and also addressing the covariate shift problem. In our implementation, we use the popular Gaussian kernel $H(X, X') \propto \exp(-\|X-X'\|^2_2/2/h^2)$, where $h$ is the bandwidth parameter \citep{hore2023conformal}. We discuss the tuning of $h$ later in Section \ref{sec_simu}.

To adjust for the second potential distributional shift between the treated and control groups, we leverage the propensity score adjustment in the classical causal inference literature \citep[see e.g.,][]{athey2015machine, shalit2017estimating, wager2018estimation, kunzel2019metalearners, farrell2021deep}, and we calculate the treatment-balancing weight as,  
\begin{equation}\label{prop}
\widehat{w}_{2} (X) = {T \over \widehat{\pi}(X)} + {1 -T \over 1- \widehat{\pi}(X)},
\end{equation}
where $\widehat{\pi}(X)$ is the estimated propensity score function using any machine learning method. In our implementation, we estimate the propensity score using the popular gradient boosting algorithm \citep{friedman2001greedy,greenwell2019gbm}. Such a weight has been shown to unbiasedly address the covariate shift due to treatment assignments \citep{chen2016personalized,zhu2020kernel,schulz2020doubly,lei2021conformal}.

Finally, we combine the two weights, and adjust the non-conformity score by the normalized weight,
\begin{eqnarray}\label{final_w}
\quad\quad \widehat{p}(X)= \frac{\widehat{w}(X)}
{\sum_{j \in \mathcal{I}_{\rm cal}\cup\{k\}}\widehat{w}(X_j)},
\end{eqnarray} 
where $\widehat{w}(X) = \widehat{w}_1(X)\widehat{w}_2(X)$.% for $j \in \mathcal{I}_{\rm cal}\cup\{k\}$.

%%%%%%%%%%%%%%%%%%%%%%%%%%%%%%%%%%%%%%%%%%%%%%%%%%%%%%%%%%%%%%%%%%%%%%%%%%%%%%%%%%%%%%%%%%%%%%%%
%\vspace{-0.1cm}
\subsection{Weighted conformal inference for ITE} 
\label{step3}
%\vspace{-0.1cm}
Based on the empirical distribution of the non-conformity scores $V_i$ by the deep generative model obtained in Section \ref{step1}, and the balancing weights $\widehat{p}(X)$ obtained in Section \ref{step2}, we obtain the conditional quantile of the non-conformity scores for the treated unit as,
\begin{eqnarray}\label{final_quant}
\widehat{Q}_{1-\alpha}\left(X_{k}\right)=\text { Quantile }_{1-\alpha}\left(\sum_{j \in \mathcal{I}_{\text{cal}}} \widehat{p}(X_j) \delta_{V_j}+\widehat{p}(X_{k}) \delta_{+\infty}\right),
\end{eqnarray}
where $\delta_s$ is the point mass at $s$, and the weights $\widehat{p}(X)$ are given in \eqref{final_w}. For a given coverage level of $1-\alpha$, we construct the predictive interval for $Y_{k}$ at the target point $X_{k}$ in $\mathcal{D}_{\rm test}$ as,
\begin{eqnarray}\label{final_CI}
\widehat{C}(X_{k}) = \cup_{m=1}^M \left[\widehat{Y}_{(k), m}-\widehat{Q}_{1-\alpha}(X_{k}), \;\; \widehat{Y}_{(k), m}+\widehat{Q}_{1-\alpha}(X_{k})\right],
\end{eqnarray}
where $\widehat{Y}_{(k),1}, \ldots, \widehat{Y}_{(k),M} $ are $M$ random samples from the estimated diffusion model $\widehat{q}_{1}(Y|X_{k})$ for the treated unit at $X_{k}$. We summarize the above procedure in Algorithm \ref{alg:1}. 

\begin{algorithm}[t!]
\caption{Conform diffusion models for individual treatment effect}
\label{alg:1}
\begin{algorithmic}[1] % The number tells where the line numbering should start
  \REQUIRE the training and calibration data $\mathcal{D}_{\rm train}, \mathcal{D}_{\rm cal}$, a testing data point $X_k$ from the testing data $\mathcal{D}_{\rm test}$, the generative sample size $M$, an estimated propensity score $\widehat{\pi}(x)$, a kernel function $H(x, x')$, and the target confidence level $\alpha$.
  \vskip 1ex  
  \STATE \underline{Step 1: Generative modeling and computing the non-conformity score.}
  \STATE Fit a generative model $\widehat{q}_1(Y|X)$ on samples from  $\mathcal{D}_{\rm train}$ with $T_i=1$.
  \STATE For each $(X_j, T_j, Y_j) \in \mathcal{D}_{\rm cal}$, sample $\widehat{Y}_{j1}, \ldots, \widehat{Y}_{jM} \sim \widehat{q}_1(Y|X_i)$.
  \STATE Calculate the non-conformity score as $V_i=\min_{1\le m \le M}|Y_i- \widehat{Y}_{im}|$.
  \vskip 1ex
  \STATE \underline{Step 2: Reweighting for covariate distributional shift.}
  \STATE For each $j \in \mathcal{D}_{\rm cal}\cup\{k\}$, calculate the weight $\widehat{w}_{1}(X_j)$ by \eqref{kernel}, and $\widehat{w}_2(X_j)$ by \eqref{prop}. 
  \STATE For each $j \in \mathcal{D}_{\rm cal}\cup\{k\}$, calculate the combined weight $\widehat{w}(X_i)=\widehat{w}_1(X_i)\widehat{w}_2(X_i)$, and the normalized weight $\widehat{p}(X_i) $ by \eqref{final_w}. 
  \vskip 1ex
  \STATE \underline{Step 3: Constructing the interval estimate.}
  \STATE Sample $\widehat{Y}_{k1}, \ldots, \widehat{Y}_{kM} \sim \widehat{q}_1(Y|X_{k})$.
  \STATE Calculate the quantile $\widehat{Q}_{1-\alpha}(X_{k})$ by \eqref{final_quant}.
  \STATE Obtain the interval estimate $\widehat{C}(X_{k})$ for $X_{k}$ by \eqref{final_CI}. 
  \ENSURE The interval estimate $\widehat{C}(X_{k})$ for $X_{k}$.
\end{algorithmic}
\end{algorithm}

%%%%%%%%%%%%%%%%%%%%%%%%%%%%%%%%%%%%%%%%%%%%%%%%%%%%%%%%%%%%%%%%%%%%%%%%%%%%%%%%%%%%%%%%%%%%%%%%
%\vspace{-0.15cm}
\section{Theory}
\label{sec_theory}
%\vspace{-0.1cm}
We next establish the coverage properties of the constructed confidence interval by our proposed method. Unlike the existing literature, our work incorporates the diffusion models and weighted conformal inference for ITE. As such, the theoretical analysis is different and highly non-trivial. Our main ideas are to leverage the techniques of conditional random sampling  \citep{pmlr-v206-wang23n} to establish the theoretical behavior of the non-conformity scores obtained by diffusion modeling, and then utilize the weighted conformal inference framework \citep{tibshirani2019conformal,lei2021conformal} to handle the discrepancy between the true weights and the estimated weights in our work. 

To begin with, we first establish the theoretical form of the weights for distributional shift adjustment. Under Assumptions \ref{cons} and \ref{ign}, the joint distribution of observed samples in the calibration data under the treated group is given by $P_{X \mid T=1} \times P_{Y(1) \mid X}$. 
We are interested in the target distribution $Q_X \times P_{Y(1) \mid X}$, where $Q_X$ is the covariate distribution in the testing data population. Following \citet{tibshirani2019conformal} and \citet{lei2021conformal}, we obtain the theoretical form of the weights as,
\begin{eqnarray*}
w_{T=1}(x) = \frac{d Q_X(x)}{d P_{X \mid T=1}(x)}=\frac{d Q_X(x)}{d P_X(x)} \frac{\mathbb{P}(T=1)}{\pi(x)}, \;
w_{T=0}(x) = \frac{d Q_X(x)}{d P_{X \mid T=0}(x)}=\frac{d Q_X(x)}{d P_X(x)} \frac{\mathbb{P}(T=0)}{1- \pi(x)}.
\end{eqnarray*}
Combining the above weights yields that, 
\begin{eqnarray*} 
w(x)= \frac{d Q_X(x)}{d P_X(x)} \left \{\frac{\mathbb{P}(T=1)}{\pi(x)}+\frac{\mathbb{P}(T=0)}{1- \pi(x)}\right \},
\end{eqnarray*}
where the first and second components can be viewed as the two weights to adjust for the two types of distributional shifts, respectively. %More details are given in Appendix Section \ref{asec_weight}. 
The coverage performance for $\widehat{C}(X_{n+1})$ in \eqref{final_CI} thus depends on the distance between the true weights $w(X)$ and the estimated weights $\widehat{w}(X)$, i.e., $\mathbb{E}_{X \sim P_X}|\widehat{w}(X)-w(X)|$. We establish and quantify the distance between the observed and target distributions through total-variation distance \citep{berrett2020conditional}. Moreover, we extend the probabilistic conformal prediction from the supervised learning \citep{pmlr-v206-wang23n} to the ITE setting. 

We make one additional assumption, which bounds both the estimated and the true weights.
\begin{assumption}\label{bnd_weights}
(Bounded Weights):
The estimated and true weights satisfy that \\
$\mathbb{E}_{X \sim P_X}\left[\widehat{w}(X) \mid  \mathcal{D}_{\text{train}}\right]<\infty$ and  $\mathbb{E}_{X \sim P_X}\left[w(X) \right]<\infty$. 
\end{assumption}
Assumption \ref{bnd_weights} is easily satisfied when Assumption \ref{posi} holds, along with a reasonable choice of a bounded kernel function. Such an assumption has been commonly imposed in the weighted conformal inference literature \citep{tibshirani2019conformal,lei2021conformal,guan2023localized}. 

We next establish the following marginal coverage property.

\begin{theorem}\label{main}
Let $n_{\rm train} = \left| \mathcal{D}_{\text{train}}\right|$, and $n_{\rm cal} = \left|\mathcal{D}_{\text{cal}}\right|$. Let $\widehat{C}(X)$ be the resulting confidence interval from Algorithm \ref{alg:1}. Then under Assumptions \ref{cons} to \ref{bnd_weights}, we have that, 
\begin{equation}\label{main_res}
\mathbb{P}_{(X, Y(1)) \sim P_X \times P_{Y(1) \mid X}}\left(Y(1) \in \widehat{C}(X)  \right) \geq 1-\alpha-\frac{1}{2} \mathbb{E}_{X \sim P_X}|\widehat{w}(X)-w(X)|.
\end{equation}
Furthermore, if $\lim_{n_{\rm train}, n_{\rm cal} \rightarrow \infty} \mathbb{E}\left| {\widehat{w} (X)}- {w(X) }\right|=0$, where $\widehat{w} (x)=\widehat{w} \left(x ;  \mathcal{D}_{\text{train}}\right)$ is an estimate of $w(X)$, then we have that, 
\begin{eqnarray}\label{main_res2}
\lim _{n_{\rm train}, n_{\rm cal} \rightarrow \infty} \mathbb{P}_{(X, Y(1)) \sim P_X \times P_{Y(1) \mid X}}\left(Y(1) \in \widehat{C}(X)  \right) \geq 1-\alpha .
\end{eqnarray}  
\end{theorem}

We give the proof of Theorem \ref{main} in Appendix Section \ref{asec_proof}. We note that, the first theoretical result in \eqref{main_res} shows that the marginal coverage probability of the proposed interval depends on the convergence rate of the estimated propensity score and the approximation error of the chosen kernel function. The second theoretical result in \eqref{main_res2} further validates the asymptotic behavior of the proposed interval, when the estimation and approximation errors go to zero.
 
We also remark that, the guarantees in Theorem \ref{main} are for the marginal coverage, and not for the conditional coverage, i.e., for a fixed $X$. However, establishing the conditional coverage is quite challenging and often involves additional much stronger assumptions \citep[see e.g., ][]{foygel2021limits,sesia2021conformal}. Moreover, the asymptotic results further require the convergence rate of the predictive model \cite[see e.g.,][]{chernozhukov2021distributional, sesia2021conformal}, which to our best knowledge is not yet available for diffusion models. Next in Section \ref{sec_simu}, we empirically evaluate the conditional coverage of our proposed confidence interval.

%%%%%%%%%%%%%%%%%%%%%%%%%%%%%%%%%%%%%%%%%%%%%%%%%%%%%%%%%%%%%%%%%%%%%%%%%%%%%%%%%%%%%%%%%%%%%%%%
%\vspace{-0.15cm}
\section{Simulation studies}
\label{sec_simu}
%\vspace{-0.15cm}
%%%%%%%%%%%%%%%%%%%%%%%%%%%%%%%%%%%%%%%%%%%%%%%%%%%%%%%%%%%%%%%%%%%%%%%%%%%%%%%%%%%%%%%%%%%%%%%%
\subsection{Simulation setup}
%\vspace{-0.1cm}
We design the simulation examples following the general settings as in \citet{lei2021conformal} but with additional complications. All numerical studies are conducted on the Wynton High-Performance Compute, with 30 parallel CPU nodes each with 4GB memory.

We generate the covariate vector $X=(X_1, \ldots, X_d)$ with $X_k = \Phi(\tilde{X}_k)$, where $\Phi$ is the cumulative distribution function (cdf) of a standard Gaussian distribution and $\tilde{X}_j$'s are independent standard Gaussian variable. We generate the potential outcome $Y(1)$ by 
\[
Y(1) = \mathbb{E}\{Y(1)|X\} + \sigma(X)\epsilon, 
\]
where the noise $\epsilon$ follows a standard Gaussian, Gamma, or a non-local moment \citep{jin2022bayesian} distributions. %We give more details on the forms of $\mathbb{E}\{Y(1)|X\}$ and the noise distributions
We give more details on the forms of $\mathbb{E}\{Y(1)|X\}$ and the noise distributions. 
Specifically, for the low-dimensional case, we set
\begin{eqnarray*}
\mathbb{E}\{Y(1)|X\} = f(X_1)f(X_2), \; \text{where} \; f(x) = \frac{2}{1+\exp\left\{-12(x-0.5)\right\} }.
\end{eqnarray*}
For the high-dimensional case, we set 
\begin{eqnarray*}
&& \mathbb{E}\{Y(1)|X\} = f_1(Z_1) f_2(Z_2) - f_3(Z_3), \;\; \text{ where } \; f_1{(x)}= \frac 2 {1+\exp\{-60(x-0.5)\}}, \\
&& 
f_2{(x)}= \frac{4}{1+ (x-0.5)^{2}}+1, \;\; 
f_3(x)= \exp\{(x-0.5)^{3}+1\}+1, \\
&&
Z_{1} = \frac{\sum_{k=1}^{d/4} w_{1k}X_{k}}{\sum_{k=1}^{d/4}w_{1k}}, \;\; 
Z_{2} = \frac{\sum_{k=d/4+1}^{d/2} w_{2k}X_{k}}{\sum_{k=d/4+1}^{d/2}w_{2k}}, \;\; 
Z_{3} = \frac{\sum_{k=d/2+1}^{d} w_{3k} X_{k}}{\sum_{k=d/2+1}^{d}w_{3k}}, \\
&& 
w_{1k}=1+\frac{9(k-1)}{d/4-1}, \;\; 
w_{2k}=1+\frac{9(k-1-d/4)}{d/4-1}, \;\; 
w_{3k}=1+\frac{9(k-1-d/2)}{d/2-1}. 
\end{eqnarray*}
For the noise, we consider a standard Gaussian distribution, a Gamma distribution with a shape parameter two and a scale parameter one, and a non-local moment noise \cite{jin2022bayesian}, with the probability density function, 
\begin{equation*}
\frac{x^{2\nu}}{C_M} \frac{1}{\sqrt{2\pi}} \exp\left(-\frac{x^2}{2}\right),
\end{equation*} 
where $C_M$ is the normalizing constant and $\nu = 5$. This distribution features two modes in its probability density function. In addition, all noises are standardized to have a standard deviation of one.

% in Appendix Section \ref{asec_simu}. 

Similar to \citet{lei2021conformal}, we set the baseline potential outcome $Y(0) = 0$. We set the propensity score $\pi(X) = 0.25 \{1+\beta_{2, 4}(X_1)\}$, where $\beta_{2, 4}$ is the cdf of the beta distribution with parameters $(2, 4)$. In our experiments, we consider two error cases, the homoscedastic case where $\sigma(X) = 1$, and the heteroscedastic case where $\sigma(X) = 0.5$ if $\mu < 0.5$ and $\sigma(X) = 5\sqrt{d/10}|\cos(\pi\mu)|$ if $\mu \ge 0.5$, and $\mu = d^{-1}\sum_{j=1}^d X_j$. For the homoscedastic case, there is no distributional shift between the calibration and testing data. For the heteroscedastic case, we consider the testing set from a subgroup of samples where $\{X: \|X\|_2 \ge q_{0.9}\}$ and $q_{0.9}$  is the $90$th quantile of $\|X\|_2$. As such, there is a clear distributional shift between the training and testing data. We also consider two covariate dimensions, the low-dimensional case where $d=10$, and the high-dimensional case where $d=300$. In the high-dimensional case, the conditional mean of $Y(1)$ is much more complicated, and relies on all covariates rather than just the first two as in the low-dimensional case. We set $n_{\rm train} = 7500, n_{\rm cal} = 2500, n_{\rm test} = 1000$, and replicate each experiment 50 times. Our goal is to construct the 95\% prediction interval for the individual treatment effect.

%%%%%%%%%%%%%%%%%%%%%%%%%%%%%%%%%%%%%%%%%%%%%%%%%%%%%%%%%%%%%%%%%%%%%%%%%%%%%%%%%%%%%%%%%%%%%%%%
%\vspace{-0.15cm}
\subsection{Comparison with alternative solutions}
%\vspace{-0.1cm}
We refer to our method as \textbf{c}onformal \textbf{d}iffusion \textbf{m}odels (CDM), and compare it with a number of alternative solutions and some variations of our proposed method, including: 
%\vspace*{-5pt}
\begin{itemize}[leftmargin=*]
\item CF: causal forest \citep{wager2018estimation}, a benchmark method to estimate CATE also compared in \cite{lei2018distribution}.
%\vspace*{-2pt}
\item CQR: conditional quantile regression \citep{lei2018distribution} using a quantile regression.
%\vspace*{-2pt}
\item CDM-nolocal: CDM without the local weights to address the shift between calibration and testing. 
%\vspace*{-2pt}
\item MLP: the multilayer perceptron method, which replaces the diffusion model with an MLP. 
%\vspace*{-2pt}
\item Naive: the naive generative method that uses the generated samples to directly obtain the interval estimate without conformal inference. 
\end{itemize}
%\vspace*{-5pt}

In our CDM implementation, we use the denoising diffusion probabilistic model as our generative model, and we adopt the neural network structure following \cite{githubcode}. %More details can be found in \texttt{./python\_scripts/} and our attached code \texttt{./mypkg/ddpm/models/ddpm\_now.py}. 
For CDM-nolocal, MLP, and the naive method, we use the same neural network architecture as CDM for a fair comparison.  
We train the neural network using the AdamW optimizer \citep{loshchilov2018decoupled}, with a weight decay of $10^{-2}$, an initial learning rate of $10^{-2}$, and a learning rate decay factor of 0.7 every 500 epochs. We choose the number of training epochs based on validation. We set the batch size for training as 128. We set the total number of diffusion steps as $400$, linearly interpolate the variance of Gaussian noise between $[10^{-4}, 0.02]$. We set the hyperparameters for CQR and CF at their default values in the corresponding \texttt{R} functions: \texttt{cfcausal::conformalCf} and \texttt{grf::causal\_forest}.

% We give more implementation details in Appendix Section \ref{asec_implement}. 

In addition, there are two key tuning parameters in CDM, the number of random samples $M$ drawn from the generative model, and the bandwidth $h$ in the kernel function $H(\cdot, \cdot)$. For the Gaussian kernel, we set $h=c\sqrt{d}$ and tune $c$. We carry out a sensitivity analysis on these tuning parameters in Section \ref{asec_sensitivity}. We have found that our method is relatively stable as long as $M$ is within a reasonable range, so we set $M=40$. We choose $c$ using a validation set that comprises 15\% of the training data. 

We employ two evaluation metrics, the empirical marginal coverage probability of ITE on the testing data,  $(1/n_{\rm test})\sum_{k=1}^{n_{\rm test}} I\{Y_k(1)-Y_k(0) \in \widehat{C}(X_k)\}$, and the median length of the constructed intervals. For a valid inference method, the empirical coverage probability should be close to or higher than $95\%$. When it does not meet this threshold, the length of interval becomes less relevant. 

\begin{figure*}[b!]
\centering
%\vspace{-0.2cm}
\includegraphics[width=1.0\textwidth, height=2.35in]{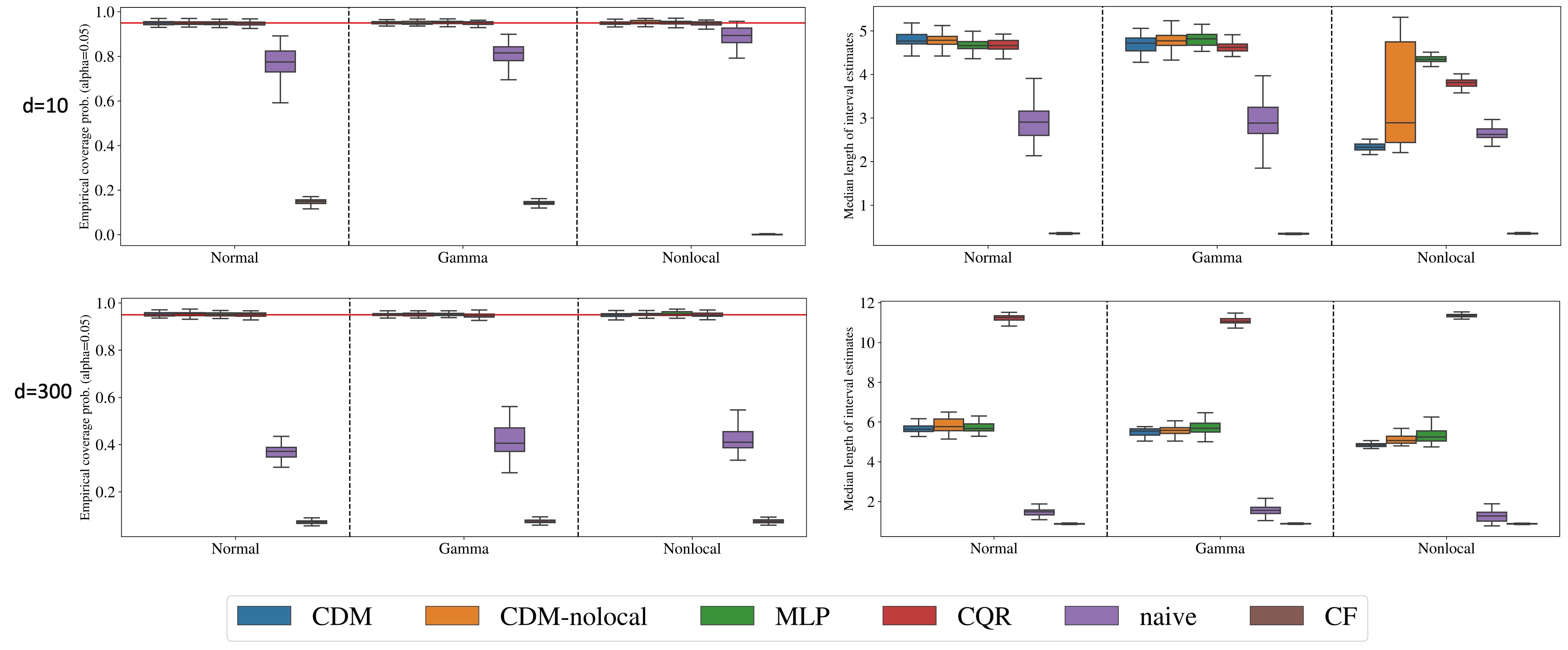}
%\vspace{-0.4cm}
\caption{Empirical coverage probability of the 95\% interval estimate for ITE (left column), and the corresponding interval length (right column) for the homoscedastic case, with a standard Gaussian, Gamma, and non-local moment noise, and the covariate dimension $d=10$ (upper row) and $d=300$ (lower row). Six methods: CDM (ours), CDM-nolocal, MLP, CQR, naive, and causal forest methods, are compared. The red horizontal line indicates the nominal coverage probability $0.95$.}
%\vspace{-0.25cm}
\label{fig:simuhomo}
\end{figure*}

Figure~\ref{fig:simuhomo} reports the results for the homoscedastic noise case when there is no distributional shift. All four conformal inference-based methods, CDM, CQR, CDM-nolocal, and MLP, show good empirical coverage regardless of noise types, whereas the naive and CF methods fail to achieve the coverage probability at the desired level. Among these four methods, when the error distribution is normal or Gamma, they achieve a similar interval length under the low-dimensional $d=10$ case. However, CDM, CDM-nolocal, and MLP clearly outperform CQR under the high-dimensional $d=300$ case. When the error has a non-local moment, the proposed CDM method achieves the shortest interval length for both the low and high-dimensional cases.
 
\begin{figure*}[t!]
\centering
%\vspace{-0.2cm}
\includegraphics[width=1.0\textwidth, height=2.35in]{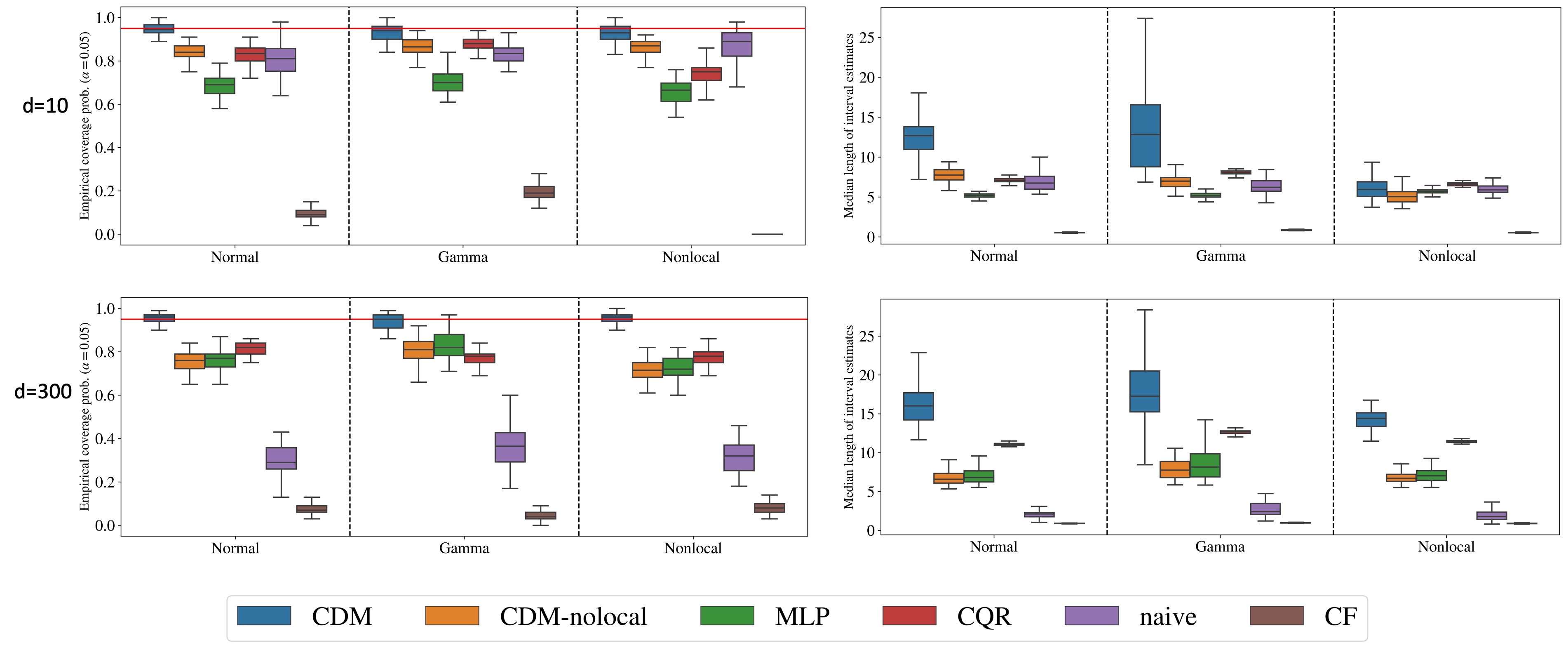}
%\vspace{-0.4cm}
\caption{Empirical coverage probability of the 95\% interval estimate for ITE (left column), and the corresponding interval length (right column) for the heteroscedastic case, with a standard Gaussian, Gamma, and non-local moment noise, and the covariate dimension $d=10$ (upper row) and $d=300$ (lower row). Six methods: CDM (ours), CDM-nolocal, MLP, CQR, naive, and causal forest methods, are compared. The red horizontal line indicates the nominal coverage probability $0.95$.}
%\vspace{-0.25cm}
\label{fig:simuheteroD1}
\end{figure*}

Figure~\ref{fig:simuheteroD1} reports the results for the heteroscedastic noise case when there is a distribution shift. Now only our proposed CDM method continues to attain the desired empirical coverage probability, while the other three conformal inference-based methods, CQR, CDM-nolocal, and MLP, all fail to do so. The CF method continues to perform poorly. The naive method shows promising results in empirical coverage probability under the low-dimensional case, but its performance declines considerably with the high-dimensional case. In addition, when the error has a non-local moment, the proposed CDM method again achieves the shortest interval length for both the low and high-dimensional cases.

%%%%%%%%%%%%%%%%%%%%%%%%%%%%%%%%%%%%%%%%%%%%%%%%%%%%%%%%%%%%%%%%%%%%%%%%%%%%%%%%%%%%%%%%%%%%%%%%
\subsection{Sensitivity analysis}
\label{asec_sensitivity}

\begin{figure}[t!]
\centering
\includegraphics[width=1.0\textwidth]{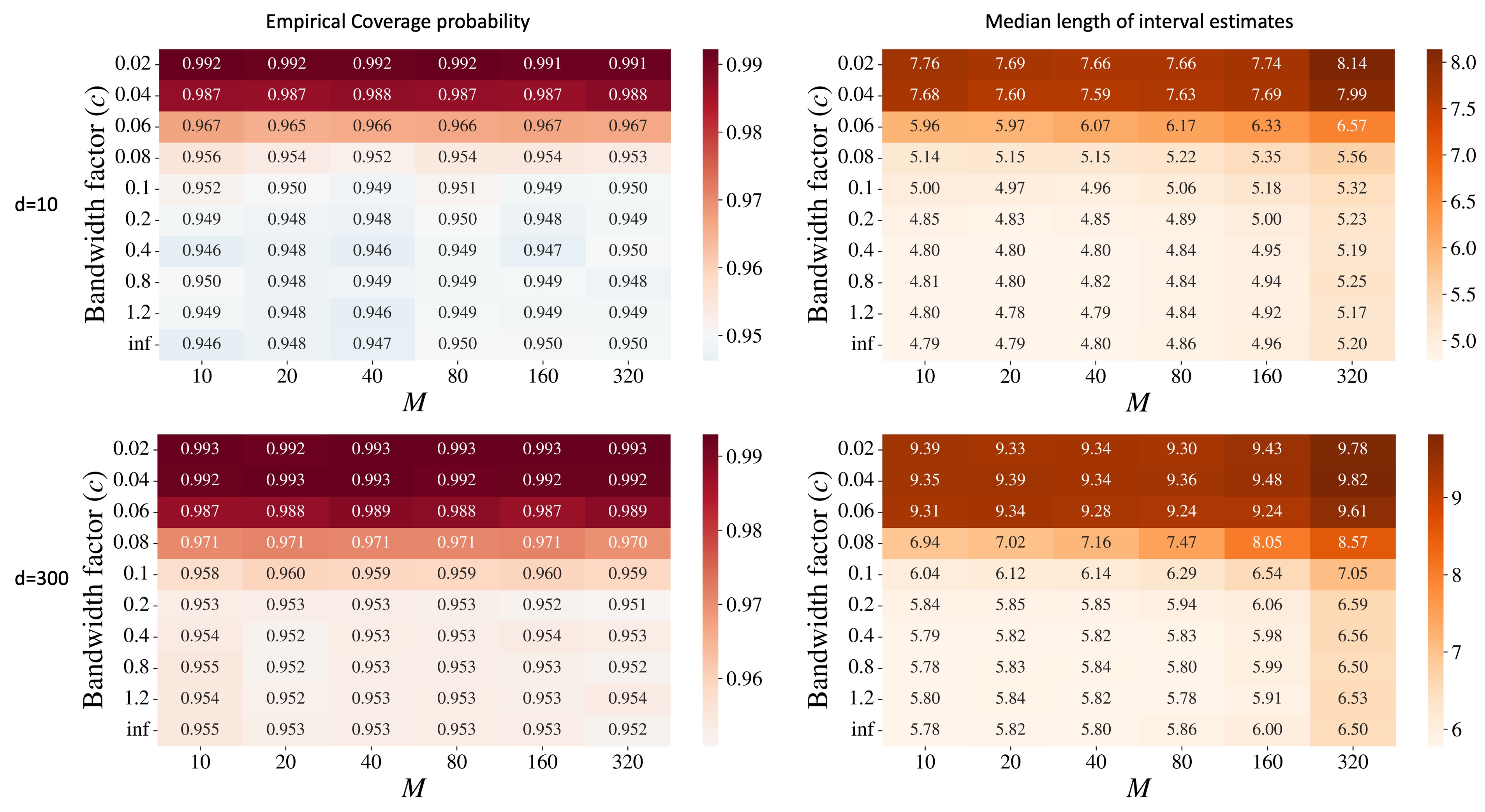}
\caption{Empirical coverage probability of the 95\% interval estimate for ITE (left column), and the corresponding interval length (right column) for the homoscedastic case, with a standard Gaussian noise, and the covariate dimension $d=10$ (upper row) and $d=300$ (lower row), under the varying number of random samples $M$ and the bandwidth factor $c$.}
\label{fig:ablhomo}
\end{figure}

\begin{figure}[t!]
\centering
\includegraphics[width=1.0\textwidth]{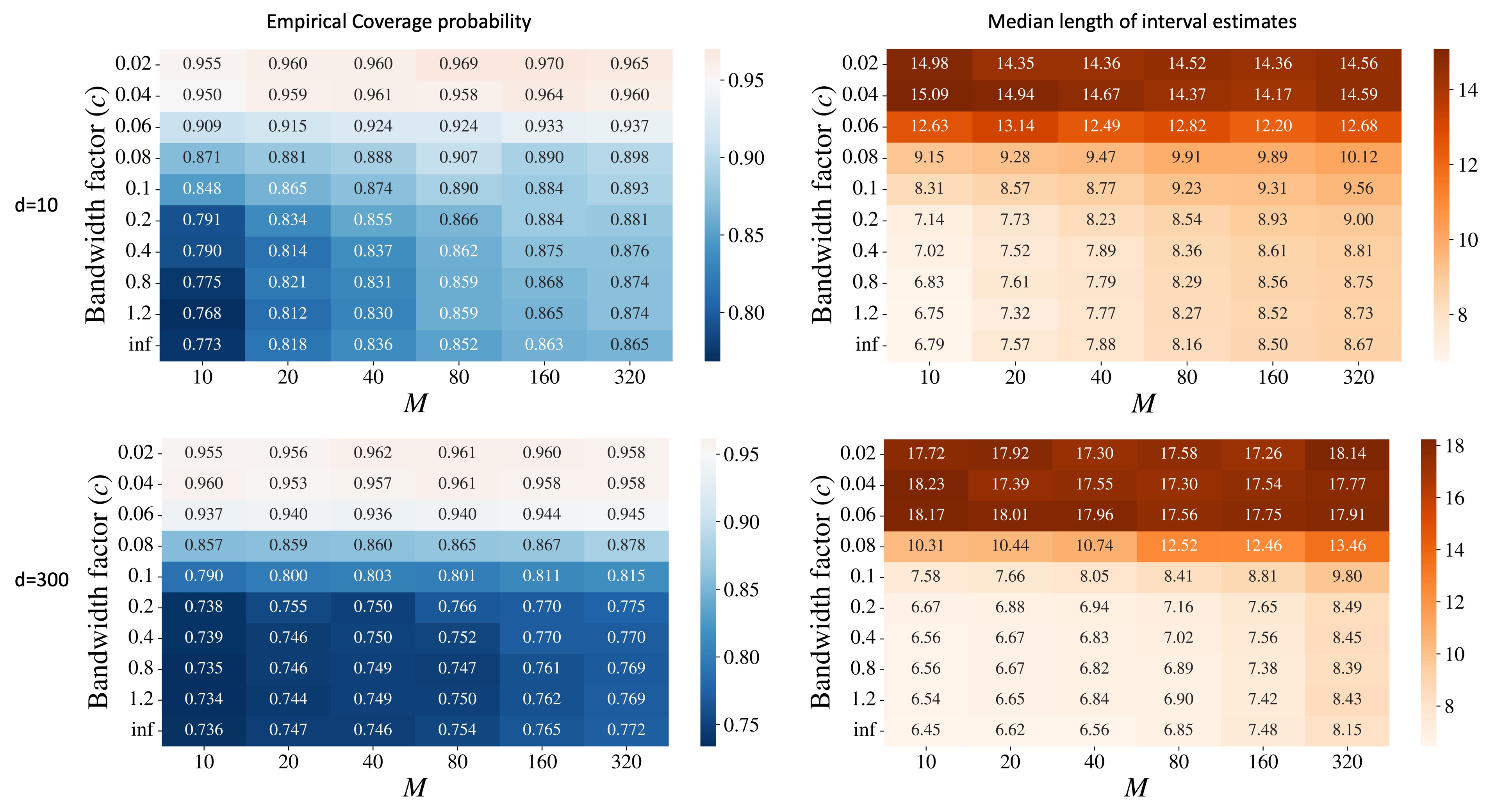}
\caption{Empirical coverage probability of the 95\% interval estimate for ITE (left column), and the corresponding interval length (right column) for the heteroscedastic case, with a standard Gaussian noise, and the covariate dimension $d=10$ (upper row) and $d=300$ (lower row), under the varying number of random samples $M$ and the bandwidth factor $c$.}
\label{fig:ablheterod1}
\end{figure}

For our proposed CDM, there are two key tuning parameters, the number of random samples $M$ drawn from the generative model, and the bandwidth $h$ in the kernel function $H(\cdot, \cdot)$. For the Gaussian kernel, we set $h=c\sqrt{d}$ and tune $c$. We next carry out a sensitivity analysis of these tuning parameters. We run the experiments for both homoscedastic and heteroscedastic cases, with $d=10$ and $300$, and a standard Gaussian noise, while we vary $K$ from $10$ to $320$, and vary $c$ from $0.02$ to $\infty$, i.e., no local approximation. 

Figures~\ref{fig:ablhomo} and \ref{fig:ablheterod1} report the results based on 50 replications. We see that our method is relatively stable as long as $M$ is within a reasonable range, so we set $M = 40$, similarly as in \citep{pmlr-v206-wang23n}. Meanwhile, we see the performance hinges on the value of $c$, so we choose $c$ using a validation set that comprises 15\% of the training data.

%%%%%%%%%%%%%%%%%%%%%%%%%%%%%%%%%%%%%%%%%%%%%%%%%%%%%%%%%%%%%%%%%%%%%%%%%%%%%%%%%%%%%%%%%%%%%%%%
%\vspace{-0.15cm}
\section{Data application}
\label{sec_real}
%\vspace{-0.15cm}
We further illustrate our method with a dataset from the electronic Intensive Care Unites (eICU) collaborative research database \citep{pollardSciData2018}, which contains over 200,000 admissions to intensive care units across the United States between 2014 and 2015. We choose the cumulative balance of metabolism as the response $Y$, the reception of IV ﬂuid resuscitation as the treatment $T$ (1 if receiving the IV ﬂuid resuscitation, and 0 if not), and we consider $d=10$ covariates, including 
age, sex, weight, Glasgow coma score, heart rate, mean blood pressure, respiratory rate, taking ventilator or not, taking vasopressor or not, and taking norepinephrine or not. The dataset contains $19716$ samples. 

We follow a similar setup as in \citet{lei2018distribution} to generate the ground truth data. Specifically, we split the data into two parts, use the first part of $7716$ samples to train models to generate the potential outcome, and use the second part of 12,000 samples to generate the covariates $X$. We apply a random forest regression on the first part of the data to obtain $\widehat{m}_1(x) = \mathbb{E}\{Y(1)|X\}$. We then fit a quantile linear regression for the $25\%$ and $75\%$ conditional quantile of $Y(1)$, and compute the corresponding conditional inter-quantile range (IQR) $\widehat{r}_1(x)$. We then generate the potential outcome $Y_i(1)$ as, 
\begin{eqnarray}\label{eq:realmodel}
Y_i(1) = \widehat{m}_1(X_i) + 0.74*\widehat{r}_1(X_i) \epsilon_i, \epsilon_i \sim \text{Gaussian}(0, 1),
\end{eqnarray}
where the factor $0.74$ is used since IQR is equal to 1.35 times the standard deviation of the Gaussian distribution. We generate the propensity score by applying a random forest model to the first part of the data, then truncate the estimated propensity score $\pi(x)$ between $0.1$ and $0.9$. We generate $n_{\rm train} = 7500$ training samples, and $n_{\rm cal} = 2500$ calibration samples from model (\ref{eq:realmodel}). We then generate an additional $1000$ sample, select those that belong to a subgroup of patients who are younger than 62, and take the norepinephrine to form the testing set, yielding $n_{\rm test} \approx 100$. We focus on this group because it is a less representative subgroup within eICU, and there is a distributional shift between the calibration and testing data. We repeat the numerical experiments $50$ times. 

\begin{table}[t!]
\centering
\caption{Empirical coverage probability of the 95\% interval estimate for ITE, and the corresponding interval length for the relatively young (age < 62) patients taking norepinephrine in the eICU data.}
\label{tab:real}
\resizebox{\textwidth}{!}{
\begin{tabular}{lcccc|lcccc}
\hline
Method 
& \multicolumn{2}{l}{Empirical probability} 
& \multicolumn{2}{l|}{Median length of interval} & 
Method 
& \multicolumn{2}{l}{Empirical probability} 
& \multicolumn{2}{l}{Median length of interval} \\
\cline{2-5}
\cline{7-10}
& Mean  & 95\% CI
& Mean  & 95\% CI & 
& Mean  & 95\% CI
& Mean  & 95\% CI \\
\hline
CDM & 0.948 & [0.938, 0.957] & 5.066 & [4.790, 5.341] & 
CQR & 0.927 & [0.920, 0.935] & 3.191 & [3.162, 3.220] \\
CDM-nolocal & 0.858 & [0.847, 0.869] & 3.092 & [3.044, 3.140] & 
CF          & 0.119 & [0.110, 0.127] & 0.242 & [0.237, 0.247] \\ 
MLP         & 0.847 & [0.836, 0.859] & 2.896 & [2.866, 2.926] &
naive       & 0.615 & [0.596, 0.635] & 1.705 & [1.656, 1.754] \\ 
\hline
\end{tabular}}
%\vspace{-0.25cm}
\end{table}

Table~\ref{tab:real} summarizes the results. Our CDM successfully provides an interval estimate with an accurate coverage probability. By contrast, the empirical coverage of CQR is lower than the target of 95\%, and that of CF is much lower. The empirical coverage of CDM-nolocal, MLP, and naive methods falls between those of CQR and CF. So none of the alternative methods provides an interval estimate with a correct coverage.

%%%%%%%%%%%%%%%%%%%%%%%%%%%%%%%%%%%%%%%%%%%%%%%%%%%%%%%%%%%%%%%%%%%%%%%%%%%%%%%%%%%%%%%%%%%%%%%%
%\vspace{-0.15cm}
\section{Conclusion and future works}\label{sec:limit}
%\vspace{-0.1cm}
In this article, we propose a conformal diffusion model-based approach to construct informative and model-free confidence intervals for individual treatment effects. Our method is versatile and can be extended to various causal effect estimation problems beyond ITE, for instance, CATE, potential outcome distribution, and decision-making. However, there are some potential limitations. Computationally, the cost is relatively high due to diffusion modeling. Combined training on different treatments may be a promising solution. Theoretically, studying the conditional coverage property beyond the marginal coverage is challenging but important, which relies on the convergence rate of the predictive model involved. We will pursue these directions in our future research.

%%%%%%%%%%%%%%%%%%%%%%%%%%%%%%%%%%%%%%%%%%%%%%%%%%%%%%%%%%%%%%%%%%%%%%%%%%%%%%%%%%%%%%%%%%%%%%%%
\newpage
\bibliography{ref}
\bibliographystyle{agsm}
% \bibliographystyle{neurips_2024}

%%%%%%%%%%%%%%%%%%%%%%%%%%%%%%%%%%%%%%%%%%%%%%%%%%%%%%%%%%%%%%%%%%%%%%%%%%%%%%%%%%%%%%%%%%%%%%%%
\newpage
\appendix

%%%%%%%%%%%%%%%%%%%%%%%%%%%%%%%%%%%%%%%%%%%%%%%%%%%%%%%%%%%%%%%%%%%%%%%%%%%%%%%%%%%%%%%%%%%%%%%%
\section{Technical proofs}\label{asec_proof}

%%%%%%%%%%%%%%%%%%%%%%%%%%%%%%%%%%%%%%%%%%%%%%%%%%%%%%%%%%%%%%%%%%%%%%%%%%%%%%%%%%%%%%%%%%%%%%%%
\subsection{Supporting lemmas}
\label{asec_lemma}

We introduce some lemmas from the literature, which will facilitate the proof of the main theorem. 

\begin{lemma}\label{lem1}
(Equation (2) in Lemma 1 from \citet{tibshirani2019conformal}). Let $v_1, \ldots, v_{n+1} \in$ $\mathbb{R}$ and $\left(p_1, \ldots, p_{n+1}\right) \in \mathbb{R}$ be non-negative real numbers summing to 1. Then, for any $\beta \in[0,1]$, 
\begin{equation*}
v_{n+1} \leq \text { Quantile }\left(\beta ; \sum_{i=1}^{n+1} p_i \delta_{v_i}\right) \Longleftrightarrow v_{n+1} \leq \text { Quantile }\left(\beta ; \sum_{i=1}^n p_i \delta_{v_i}+p_{n+1} \delta_{\infty}\right) .
\end{equation*}
\end{lemma}

\begin{lemma}
(Equation (10) from \citet{berrett2020conditional}). Let $d_{\mathrm{TV}}\left(Q_{1 X}, Q_{2 X}\right)$ denote the total-variation distance between two distributions $Q_{1 X}$ and $Q_{2 X}$. Then, 
\begin{equation*}
d_{\mathrm{TV}}\left(Q_{1 X} \times P_{Y \mid X}, Q_{2 X} \times P_{Y \mid X}\right)=d_{\mathrm{TV}}\left(Q_{1 X}, Q_{2 X}\right) .
\end{equation*}
\end{lemma}

%%%%%%%%%%%%%%%%%%%%%%%%%%%%%%%%%%%%%%%%%%%%%%%%%%%%%%%%%%%%%%%%%%%%%%%%%%%%%%%%%%%%%%%%%%%%%%%%
\subsection{Proof of Theorem \ref{main}}

We first recall the following notations. We divide the observed triplets $(Y_i, T_i, X_i)_{i=1}^n$ into three parts: $\mathcal{D}_{\text{train}}={(X_i, T_i, Y_i)}_{i=1}^{n_{\text{train}}}$ as the training data, $\mathcal{D}_{\text{cal}}={(X_j, T_j, Y_j)}_{j=1}^{n_{\text{cal}}}$ as the calibration data, and $\mathcal{D}_{\text{test}}={(X_k, T_k, Y_k)}_{k=1}^{n_{\text{test}}}$ as the holdout testing or target data. Let $\mathcal{I}_{\text{train}}, \mathcal{I}_{\text{cal}}, \mathcal{I}_{\text{test}}$ denote the corresponding index set, respectively. Without loss of generality, we follow the general literature of ITE \citep[see e.g., ][]{lei2021conformal} and let $Y(0) = 0$. Then the confidence interval for $\tau_i$  is equivalent to that for $Y_i(1)$. Define $\pi(x) = \mathbb{P}(T=1 \mid X=x)$ as the propensity score function that describes the treatment assignment mechanism. 

We propose to estimate each component of the weight $w(X) = w_1(X) w_2(X)$ as follows. We estimate $w_1(X)$ as
\begin{eqnarray*}
\widehat{w}_{1}(X)=\frac{H\left(X, \widetilde{X}_{k}\right)}{\sum_{j \in \mathcal{I}_{\rm cal}\cup\{k\}} H\left(X_j, \widetilde{X}_{k}\right)}, 
\end{eqnarray*}
%{\color{blue} 
where $\widetilde{X}_{k}$ is sampled from the density function proportional to $H(X_{k}, \cdot)$. We estimate $w_2(X)$ as
\begin{equation*}
\widehat{w}_{2} (X) = {T \over \widehat{\pi}(X)} + {1 -T \over 1- \widehat{\pi}(X)},
\end{equation*}
where $\widehat{\pi}(X)$ is the estimated propensity score using any machine learning method. The final weight to simultaneously address two covariate shifts is $\widehat{w}(X) = \widehat{w}_1(X)\widehat{w}_2(X)$.

Next, we establish and quantify the distance between the observed and target distributions through the distance between the estimated $\widehat{w}(X)$ and the true weights ${w} (X)$. We consider the case where $Q_X$ is absolutely continuous with respect to $P_X$, i.e., $\mathbb{P}_{X \sim Q_X}(w(X)<\infty)=1$. The case when $\mathbb{P}_{X \sim Q_X}(w(X)<\infty)<1$ can be proved similarly following the proof of Theorem 3 in \citet{lei2021conformal}. 

For any measurable function $f$, 
\begin{equation}\label{true_w}
\mathbb{E}_{X \sim Q_X}[f(X)]=\mathbb{E}_{X \sim P_X}[w(X) f(X)].
\end{equation}
Meanwhile, it always holds that $\mathbb{P}_{X \sim P_X}(w(X)<\infty)=1$. In addition, the assumption $\mathbb{E}_{X \sim P_X}\left[\widehat{w}(X) \mid  \mathcal{D}_{\text{train}}\right]<\infty$ implies that $\mathbb{P}_{X \sim P_X}(\widehat{w}(X)<\infty)=1$. By \eqref{true_w},
\begin{equation*}
\mathbb{P}_{X \sim Q_X}(\widehat{w}(X)<\infty)=1-\mathbb{E}_{X \sim P_X}[w(X) I(\widehat{w}(X)=\infty)].
\end{equation*}
Since the integrand is non-negative,
\begin{equation*}
\begin{aligned}
& \mathbb{E}_{X \sim P_X}[w(X) I(\widehat{w}(X)=\infty)] \\
& \quad=\lim _{K \rightarrow \infty} \mathbb{E}_{X \sim P_X}[w(X) I(w(X) \leq K, \widehat{w}(X)=\infty)] \leq \lim _{K \rightarrow \infty} K \mathbb{P}_{X \sim P_X}(\widehat{w}(X)=\infty)=0 .
\end{aligned}
\end{equation*}
As such, we also have $ \mathbb{P}_{X \sim Q_X}(\widehat{w}(X)<\infty)=1$.

Let $\left(X_{k}, Y_{k}\right) \sim Q_X \times P_{Y \mid X}$ for $k \in \mathcal{I}_{\rm test}$ as the target sample of interest. Write $Z_i$ for $\left(X_i, Y_i\right)$ and $V$ for $\left(V_{j}\right)_{j \in \mathcal{I}_{\rm cal}\cup\{k\}}$.  Next, for any permutation $\pi$ on $\{\mathcal{I}_{\rm cal}\cup\{k\}\}$ and $v^* \in$ $\mathbb{R}^{n_{\text{cal}}+1}$, let $v_\pi^*=\left(v_{\pi(1)}^*, \ldots, v_{\pi(n_{\text{cal}}+1)}^*\right)$. Furthermore, let $\ell(z)$ be the joint density of $\mathcal{Z}=\left(Z_1, \ldots, Z_{n_{\text{cal}}+1}\right)$, and $p(z)$ be the density of $Z_1$ with respect to a dominating measure. Letting $\mathcal{E}(v)$ denote the unordered set of $v$, then, 
\begin{equation*}
\left(V \mid \mathcal{E}(V)=\mathcal{E}\left(v^*\right),  \mathcal{D}_{\text{train}}\right) \stackrel{d}{=} v_{\Pi}^*,
\end{equation*}
where $\Pi$ is a random permutation with
\begin{equation*}
\mathbb{P}\left(\Pi=\pi \mid  \mathcal{D}_{\text{train}}\right)=\frac{p\left(z_\pi^*\right)}{\sum_\pi p\left(z_\pi^*\right)}=\frac{w\left(X_{\pi(n_{\text{cal}}+1)}\right)}{\sum_\pi w\left(X_{\pi(n_{\text{cal}}+1)}\right)}=\frac{w\left(X_{\pi(n_{\text{cal}}+1)}\right)}{n!\sum_{i \in \mathcal{I}_{\rm cal}\cup\{k\}} w\left(X_i\right)} .
\end{equation*}
Note that this conditional probability is well-defined, because $w(X)<\infty$ almost surely under both $P_X$ and $Q_X$. As a result, for any $j \in \mathcal{I}_{\rm cal}\cup\{k\}$,
\begin{equation*}
\mathbb{P}\left(\Pi(n_{\text{cal}}+1)=j \mid  \mathcal{D}_{\text{train}}\right)=\frac{w\left(X_j\right)}{\sum_{i \in \mathcal{I}_{\rm cal}\cup\{k\}} w\left(X_i\right)}=p\left(X_j\right).
\end{equation*}
% where $p_{n+1}$ denotes $p_{\infty}$ for notational convenience. 
This gives that, 
\begin{equation}\label{Vnp1}
\left(V_{k} \mid \mathcal{E}(V)=\mathcal{E}\left(v^*\right),  \mathcal{D}_{\text{train}}\right) \stackrel{d}{=} v_{\Pi(n_{\text{cal}}+1)}^* \sim \sum_{j \in \mathcal{I}_{\rm cal}\cup\{k\}} p\left(X_{j}\right) \delta_{v_j^*} .
\end{equation}
Note that $p(x)$ involves the true likelihood ratio function $w(x)$, and thus is different from $\widehat{p}(x)$. Let $\tilde{Q}_X$ be a measure with
\begin{equation*}
d \tilde{Q}_X(x)=\widehat{w}(x) d P_X(x) .
\end{equation*}
Since $\mathbb{E}_{X \sim P_X}[\widehat{w}(X)]=1, \mathbb{P}_{X \sim P_X}(\widehat{w}(X)<\infty)=1$. As a result, $\tilde{Q}_X$ is a probability measure. Consider now a new sample $\left(\tilde{X}_{k}, \tilde{Y}_{k}\right) \sim \tilde{Q}_X \times P_{Y \mid X}$. Let $\tilde{V}_{k}$ denote the non-conformity score of $\left(\tilde{X}_{k}, \tilde{Y}_{k}\right)$, and set $\tilde{V}=\left(V_{j}\right)_{j \in \mathcal{I}_{\rm cal}}\cup \tilde{V}_{k} $. Using the same argument as for \eqref{Vnp1}, we have that 
\begin{equation*}
\left(\tilde{V}_{k} \mid \mathcal{E}(\tilde{V})=\mathcal{E}\left(v^*\right),  \mathcal{D}_{\text{train}}\right) \sim \sum_{j \in \mathcal{I}_{\rm cal}\cup\{k\}} \widehat{p}\left(\tilde{X}_{j}\right) \delta_{v_j^*} .
\end{equation*}
Note that each $\widehat{p}\left(\tilde{X}_{j}\right)$ for $j \in \mathcal{I}_{\rm cal}\cup\{k\}$ is well-defined, because $\widehat{w}\left(X_j\right)$ is almost surely finite under both $P_X$ and $Q_X$. 

For a target coverage level of $1-\alpha$, we construct the prediction set for $Y_{k}(1)$ in  \eqref{final_CI} as
 \begin{eqnarray*} 
\widehat{C}(X_{k}) = \cup_{m=1}^M \left[\widehat{Y}_{(k), m}-\widehat{Q}_{1-\alpha}(X_{k}), \;\; \widehat{Y}_{(k), m}+\widehat{Q}_{1-\alpha}(X_{k})\right],
\end{eqnarray*}
which can be rewritten as 
\begin{equation*}
\widehat{C}(X_{k})=\cup_{m=1}^M\left\{y:\left\|y-\widehat{Y}_{(k), m}\right\| \leq \widehat{Q}_{1-\alpha}(X_{k})\right\},
\end{equation*}
where 
\begin{eqnarray*}
\widehat{Q}_{1-\alpha}\left(X_{k}\right)=\text { Quantile }_{1-\alpha}\left(\sum_{j \in \mathcal{I}_{\text{cal}}} \widehat{p}(X_j) \delta_{V_j}+\widehat{p}(X_{k}) \delta_{+\infty}\right),
\end{eqnarray*}
and $\delta_s$ is the point mass at $s$, and the weights are given by \eqref{final_w}. 

Next, we prove the fact, following \citet{pmlr-v206-wang23n}, that 
\begin{eqnarray}\label{eqv_tras}
Y_{k}(1) \in \widehat{C}\left(X_{k}\right) \Leftrightarrow V_{k} \leq \widehat{Q}_{1-\alpha}\left(X_{k}\right),
\end{eqnarray}
where $V_{k}= \min_{1\le m \le M}|Y_k- \widehat{Y}_{(k),m}|$. 
Suppose the left-hand-side in \eqref{eqv_tras} is true, then $\exists m, 1 \leq m \leq M$, such that,
\begin{eqnarray*}
Y_{k}(1) \in\left\{y:\left\|y-\widehat{Y}_{(k), m}\right\| \leq \widehat{Q}_{1-\alpha}\left(X_{k}\right)\right\}.
\end{eqnarray*}
This means $\left\|Y_{k}(1)-\widehat{Y}_{(k), m}\right\| \leq \widehat{Q}_{1-\alpha}\left(X_{k}\right)$. Therefore, we have that
\begin{eqnarray*}
V_{k}=\min_m\left\|Y_{k}(1)-\widehat{Y}_{(k), m}\right\| \leq\left\|Y_{k}(1)-\widehat{Y}_{(k), m}\right\| \leq \widehat{Q}_{1-\alpha}\left(X_{k}\right).
\end{eqnarray*}

On the other hand, suppose the right-hand-side in \eqref{eqv_tras} is true. Denote 
\begin{eqnarray*}
t=\arg \min _m\left\|Y_{k}(1)-\widehat{Y}_{(k), m}\right\|.
\end{eqnarray*}
Then we have $\left\|Y_{k}(1)-\widehat{Y}_{(k), t}\right\| \leq \widehat{Q}_{1-\alpha}\left(X_{k}\right)$, i.e., 
\begin{eqnarray*}
Y_{k}(1) \in\left\{y:\left\|y-\widehat{Y}_{(k), t}\right\| \leq \widehat{Q}_{1-\alpha}\left(X_{k}\right)\right\}.
\end{eqnarray*}
Therefore, we have that $Y_{k}(1) \in \widehat{C}\left(X_{k} \right)$.

As a consequence, we have that
\begin{eqnarray}\label{mid1}
\begin{aligned}
&\mathbb{P}\left(\tilde{Y}_{k} \in \widehat{C}\left(\tilde{X}_{k}\right) \mid  \mathcal{D}_{\text{train}}\right) =\mathbb{P}\left(\tilde{V}_{k} \leq \widehat{Q}_{1-\alpha}\left(\tilde{X}_{k}\right) \mid  \mathcal{D}_{\text{train}}\right) \\
= \; &\mathbb{P}\left(\tilde{V}_{k} \leq \text { Quantile }_{1-\alpha}\left(\sum_{i \in \mathcal{I}_{\text{cal}}} \widehat{p}  \left(\tilde{X}_{i}\right) \delta_{V_i}+\widehat{p} \left(\tilde{X}_{k}\right) \delta_{+\infty}\right)
  \mid  \mathcal{D}_{\text{train}}\right) 
\end{aligned}
\end{eqnarray}
Furthermore, applying Lemma \ref{lem1}, we have that
\begin{eqnarray}\label{mid2}
\begin{aligned}
&\mathbb{P}\left(\tilde{V}_{k} \leq \text { Quantile }_{1-\alpha}\left(\sum_{i \in \mathcal{I}_{\text{cal}}} \widehat{p} \left(\tilde{X}_{i}\right) \delta_{V_i}+\widehat{p} \left(\tilde{X}_{k}\right) \delta_{+\infty}\right)
  \mid  \mathcal{D}_{\text{train}}\right) \\
= \; &\mathbb{P}\left(\tilde{V}_{k} \leq \text { Quantile }_{1-\alpha}\left(\sum_{i \in \mathcal{I}_{\text{cal}}} \widehat{p} \left(\tilde{X}_{i}\right) \delta_{V_i}+\widehat{p} \left(\tilde{X}_{k}\right) \delta_{\tilde{V}_{k}}\right) \mid  \mathcal{D}_{\text{train}}\right) \\
= \; &\mathbb{E} \mathbb{P}\left(\tilde{V}_{k} \leq\text { Quantile }_{1-\alpha}\left(\sum_{i \in \mathcal{I}_{\text{cal}}} \widehat{p} \left(\tilde{X}_{i}\right) \delta_{V_i}+\widehat{p} \left(\tilde{X}_{k}\right) \delta_{\tilde{V}_{k}}\right) \mid \mathcal{E}(\tilde{V}),  \mathcal{D}_{\text{train}}\right)\equiv \eta.
\end{aligned}
\end{eqnarray}
Finally, following \citet{lei2021conformal}, recalling the definition of $\tilde{V}_{k}$ and the quantile function, we have that $\eta \geq 1-\alpha$. Combining it with \eqref{mid1} and \eqref{mid2}, we have that
\begin{eqnarray}\label{res1}
\mathbb{P}\left(\tilde{Y}_{k} \in \widehat{C}\left(\tilde{X}_{k}\right) \mid  \mathcal{D}_{\text{train}}\right) =\mathbb{P}\left(\tilde{V}_{k} \leq \widehat{Q}_{1-\alpha}\left(\tilde{X}_{k}\right) \mid  \mathcal{D}_{\text{train}}\right) \geq 1-\alpha.
\end{eqnarray}

Applying Lemma 2, we obtain that
\begin{eqnarray*}
d_{\mathrm{TV}}\left(Q_X \times P_{Y \mid X}, \tilde{Q}_X \times P_{Y \mid X}\right)=d_{\mathrm{TV}}\left(Q_X, \tilde{Q}_X\right).
\end{eqnarray*}
As a consequence,
\begin{eqnarray*}
\left|\mathbb{P}\left(Y_{k}(1) \in \widehat{C}\left(X_{k}\right) \mid  \mathcal{D}_{\text{train}}, \mathcal{D}_{\text{cal}}\right)-\mathbb{P}\left(\tilde{Y}_{k} \in \widehat{C}\left(\tilde{X}_{k}\right) \mid  \mathcal{D}_{\text{train}}, \mathcal{D}_{\text{cal}}\right)\right| \leq d_{\mathrm{TV}}\left(Q_X, \tilde{Q}_X\right),
\end{eqnarray*}
which implies that
  \begin{eqnarray*}
\mathbb{P}\left(Y_{k}(1) \in \widehat{C}\left(X_{k}\right) \mid  \mathcal{D}_{\text{train}}, \mathcal{D}_{\text{cal}}\right) \geq \mathbb{P}\left(\tilde{Y}_{k} \in \widehat{C}\left(\tilde{X}_{k}\right) \mid  \mathcal{D}_{\text{train}}, \mathcal{D}_{\text{cal}}\right)-d_{\mathrm{TV}}\left(Q_X, \tilde{Q}_X\right) .
  \end{eqnarray*}
Taking expectation over $\mathcal{D}_{\text{cal}}$, and combining with the results from \eqref{res1}, we have that
\begin{eqnarray*}
\begin{aligned}
&\mathbb{P}\left(Y_{k}(1) \in \widehat{C}\left(X_{k}\right) \mid  \mathcal{D}_{\text{train}}\right) \\
\geq &\mathbb{P}\left(\tilde{Y}_{k} \in \widehat{C}\left(\tilde{X}_{k}\right) \mid  \mathcal{D}_{\text{train}}\right)-d_{\mathrm{TV}}\left(Q_X, \tilde{Q}_X\right) \geq 1-\alpha-d_{\mathrm{TV}}\left(Q_X, \tilde{Q}_X\right).
\end{aligned}
\end{eqnarray*}

Using the integral definition of total-variation distance and \eqref{true_w},
\begin{eqnarray*}
\begin{aligned}
d_{\mathrm{TV}}\left(Q_X, \tilde{Q}_X\right) & =\frac{1}{2} \int\left|\widehat{w}(x) d P_X(x)-d Q_X(x)\right|=\frac{1}{2} \int\left|\widehat{w}(x) d P_X(x)-w(x) d P_X(x)\right| \\
& =\frac{1}{2} \mathbb{E}_{X \sim P_X}|\widehat{w}(X)-w(X)|.
\end{aligned}
\end{eqnarray*} 

Finally, taking expectation over $ \mathcal{D}_{\text{train}}$, we obtain that 
\begin{eqnarray}\label{main_thm1} 
\mathbb{P}_{(X, Y(1)) \sim P_X \times P_{Y(1) \mid X}}\left(Y(1) \in \widehat{C}(X)  \right) \geq 1-\alpha-\frac{1}{2} \mathbb{E}_{X \sim P_X}|\widehat{w}(X)-w(X)|. 
\end{eqnarray} 
This proves the result in \eqref{main_res} of Theorem \ref{main}.

Furthermore, when $\lim_{n_{\rm train}, n_{\rm cal} \rightarrow \infty} \mathbb{E}\left| {\widehat{w} (X)}- {w(X) }\right|=0$, we have that, 
\begin{eqnarray*}
\lim _{n_{\rm train}, n_{\rm cal} \rightarrow \infty} \mathbb{P}_{(X, Y(1)) \sim P_X \times P_{Y(1) \mid X}}\left(Y(1) \in \widehat{C}(X)  \right) \geq 1-\alpha .
\end{eqnarray*}  
This proves the result in \eqref{main_res2} of Theorem \ref{main}.

Together this completes the proof of Theorem \ref{main}.

\end{document}